\documentclass[acmtog,screen,acmauthoryear,timestamp]{acmart}
\setcitestyle{square}

\usepackage{booktabs} 
\usepackage{float}

\setcopyright{rightsretained}

\acmDOI{10.475/123_4}

\acmISBN{123-4567-24-567/08/06}

\acmConference[SIGGRAPH ASIA'18]{ACM conference}{July 2018}{El
  Paso, Texas USA} 
\acmYear{2018}
\copyrightyear{2018}
\acmArticle{282}
\acmSubmissionID{282}
\acmPrice{15.00}

\DeclareMathOperator*{\argmin}{argmin}

\begin{document}

\title{Neural Rendering and Reenactment of Human Actor Videos}

\begin{teaserfigure}
\rotatebox[origin=l]{90}{\hspace{2mm}\textbf{Output} (synth.) \hspace{11mm} \textbf{Driving motion} (real)}
	\includegraphics[width=\textwidth]{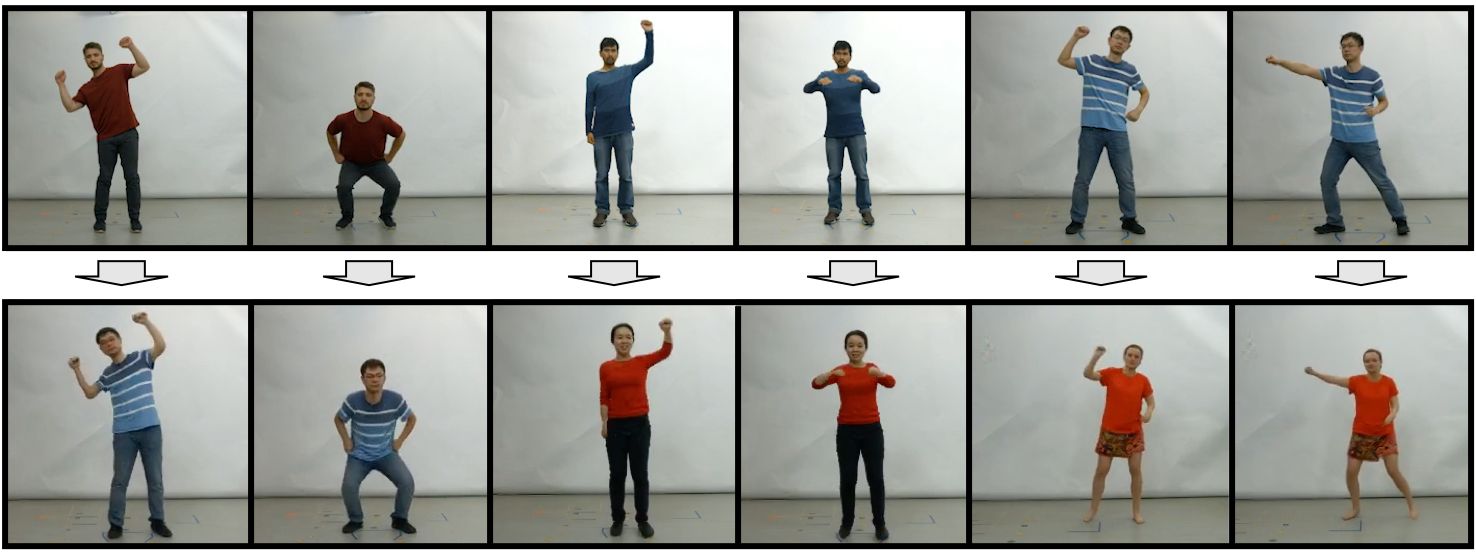}
    \vspace{-6mm}
	\caption{We propose a novel learning-based approach for the animation and reenactment of human actor videos. The top row shows some frames of the video from which the \emph{source motion} is extracted, and the bottom row shows the corresponding synthesized target person imagery reenacting the source motion.}
	\label{fig:teaser}
\end{teaserfigure}

\author{Lingjie Liu}
\email{liulingjie0206@gmail.com}
\affiliation{%
	\institution{University of Hong Kong, Max Planck Institute for Informatics}
}

\author{Weipeng Xu}
\email{wxu@mpi-inf.mpg.de}
\affiliation{%
	\institution{Max Planck Institute for Informatics}
}

\author{Michael Zollh\"ofer}
\email{zollhoefer@cs.stanford.edu}
\affiliation{%
	\institution{Stanford University, Max Planck Institute for Informatics}
}
\author{Hyeongwoo Kim}
\email{hyeongwoo.kim@mpi-inf.mpg.de}

\author{Florian Bernard}
\email{fbernard@mpi-inf.mpg.de}

\author{Marc Habermann}
\email{mhaberma@mpi-inf.mpg.de}
\affiliation{%
	\institution{Max Planck Institute for Informatics}
}

\author{Wenping Wang}
\email{wenping@cs.hku.hk}
\affiliation{%
	\institution{University of Hong Kong}
}

\author{Christian Theobalt}
\email{theobalt@mpi-inf.mpg.de}
\affiliation{%
	\institution{Max Planck Institute for Informatics}
}

\begin{abstract}
We propose a method for generating video-realistic animations of real humans under user control. In contrast to conventional human character rendering, we do not require the availability of a production-quality photo-realistic 3D model of the human, but instead rely on a video sequence in conjunction with a (medium-quality) controllable 3D template model of the person. With that, our approach significantly reduces production cost compared to conventional rendering approaches based on production-quality 3D models, and can also be used to realistically edit existing videos. Technically, this is achieved by training a neural network that translates simple synthetic images of a human character into realistic imagery. For training our networks, we first track the 3D motion of the person in the video using the template model, and subsequently generate a synthetically rendered version of the video. These images are then used to train a conditional generative adversarial network that translates synthetic images of the 3D model into realistic imagery of the human. We evaluate our method for the reenactment of another person that is tracked in order to obtain the motion data, and show video results generated from artist-designed skeleton motion. Our results outperform the state-of-the-art in learning-based human image synthesis.
	
\end{abstract}



\maketitle

\section{Introduction}
The creation of realistically rendered and controllable animations of human characters is a crucial task in many computer graphics applications. 
Virtual actors play a key role in games and visual effects, in telepresence, or in virtual and augmented reality.
Today, the plausible rendition of video-realistic characters---i.e., animations indistinguishable from a video of a human---under user control is also important in other domains, such as in simulation environments that render training data of the human world for camera-based perception algorithms of autonomous systems and vehicles~\cite{Dosovitskiy17}. 
There, simulated characters can enact large corpora of annotated real world scenes and actions, which would be hard to actually capture in the real world.
Also, training data for dangerous situations, like a child running unexpectedly onto a street, cannot be captured in reality, but such image data are crucial for training of autonomous systems.

With established computer graphics modeling and rendering tools, creating a photo-real virtual clone of a real human that is indistinguishable from a video of the person is still a highly complex and time consuming process. 
To achieve this goal, high quality human shape and appearance models need to be hand-designed or captured from real individuals with sophisticated scanners. 
Real world motion and performance data needs to be artist-designed or captured with dense camera arrays, and sophisticated global illumination rendering methods are required to display animations photo-realistically. 
In consequence, creation and rendering of a video-realistic virtual human is highly time consuming.

We therefore propose a new efficient and lightweight approach to capture and render video-realistic animations of real humans under user control.
At runtime it requires only a monocular color video of a person as input (or any other motion source) that is then used to control the animation of a reenacted video of a different actor. 
In order to achieve this, we employ a learning-based approach that renders realistic human images merely based on synthetic human animations.

At the training stage, our method takes two short monocular videos of a person as input, one in static posture, and one in general motion.
From the \emph{static} posture video, a fully textured 3D surface model of the actor with a rigged skeleton is reconstructed.
This character model is then used to capture the skeletal motion seen in the \emph{motion} video using (a modified version of) the monocular human performance capture method of ~\cite{VNect_SIGGRAPH2017}. 
While this captures the 3D pose and surface motion of the actor, there is still a significant gap to
the expected photo-realistic appearance of the virtual character.
Hence, we train a generative neural network in an attempt to fill this gap.
Specifically, based on the 3D character model and the tracked motion data, we first render out different image modalities of the animated character (color and depth images, body part segmentations), which correspond to the image frames in the motion video.
Then, based on the so-created training data, we train a conditional GAN to reconstruct photo-realistic imagery of the motion video frames using our rendered images as conditioning input.

During testing, we animate the virtual 3D character of the target subject with a user-defined motion sequence, which can stem from an arbitrary source (e.g. motion capture (MoCap) data, artist-designed animations, or videos of an actor), and then render the color image, depth, and semantic masks for each frame of the output video of the virtual character.
Then, we pass the rendered conditioning images to the network and thus obtain photo-realistic video of the same person performing the desired motion.

We emphasize that, compared to previous work that mapped face model renderings to realistic face video (\emph{Deep Video Portraits} \cite{kim2018DeepVideo}), translating complete articulated character renderings to video is a much more difficult problem due to more severe pose and appearance changes and the so-resulting non-linearities and discontinuities. Another difficulty is the inevitable imperfection in human body tracking, which directly results in a misalignment between the conditioning input and the ground truth image.
Hence, established image-to-image translation approaches like \emph{pix2pix}~\cite{IsolaZZE2017} and \emph{Deep Video Portraits} are not directly applicable to full human body performances. 
To alleviate these problems, we propose a novel GAN architecture that is based on two main contributions: 
\begin{itemize}
	\item[(i)] a part-based dense rendering of the model in RGB and depth channel as conditioning images to better constrain the pose-to-image translation problem and disambiguate highly articulated motions, and
	\item[(ii)] an attentive discriminator network tailored to the character translation task that enforces the network to pay more attention to regions where the image quality is still low.
\end{itemize} 
The proposed method allows us to reenact humans in video by using driving motion data from arbitrary sources and to synthesize video-realistic target videos.
In our experiments, we show high quality video reenactment and animation results on several challenging sequences, and show clear improvements over most related previous work. 

\section{Related work}

We focus our discussion on the most related performance capture, video-based rendering, and generative modeling approaches.

\paragraph{Video-based Characters and Free-viewpoint Video}

Video-based synthesis tries to close the gap between photo-realistic videos and rendered controllable characters.
First approaches were based on reordering existing video clips \cite{Schodl:2002,Schodl:2000VT}.
Recent techniques enable video-based characters \cite{Xu:SIGGRPAH:2011,Li:2017,Casas:2014,Volino2014} and free-viewpoint video \cite{Carranza:2003,Li:2014,zitnick2004high,Collet:2015} based on 3D proxies.
Approaches for video-based characters either use dynamically textured meshes and/or image-based rendering techniques.
Dynamic textures \cite{Casas:2014,Volino2014} can be computed by temporal registration of multi-view footage in texture space.
The result is fully controllable, but unfortunately the silhouettes of the person match the coarse geometric proxy.
The approach of \citet{Xu:SIGGRPAH:2011} synthesizes plausible videos of a human actors with new body motions and viewpoints.
Synthesis is performed by finding a coherent sequence of frames that matches the specified target motion and warping these to the novel viewpoint.
\citet{Li:2017} proposed an approach for sparse photo-realistic animation based on a single RGBD sensor.
The approach uses model-guided texture synthesis based on weighted low-rank matrix completion.
Most approaches for video-based characters require complex and controlled setups, have a high runtime, or do not generalize to challenging motions.
In contrast, our approach is based on a single consumer-grade sensor, is efficient and generalizes well.

\paragraph{Learned Image-to-image Translation}

Many problems in computer vision and graphics can be phrased as image-to-image mappings.
Recently, many approaches employ convolutional neural networks (CNNs) to learn the best mapping based on large training corpora.
Techniques can be categorized into (variational) auto-encoders (VAEs) \cite{HintoS2006,jKingma2014}, autoregressive models (AMs) \cite{Oord:2016}, and conditional generative adversarial networks (cGANs) \cite{GoodfPMXWOCB2014,RadfoMC2016,MirzaO2014,IsolaZZE2017}.
Encoder-decoder architectures \cite{HintoS2006} are often used in combination with skip connections to enable feature propagation on different scales.
This is similar to U-Net \cite{RonneFB2015}.
CGANs \cite{MirzaO2014,IsolaZZE2017} have obtained impressive results on a wide range of tasks.
Recently, high-resolution image generation has been demonstrated based on cascaded refinement networks \cite{ChenK2017} and progressively trained (conditional) GANs \cite{KarraALL2018,WangLZTKC2018}.
These approaches are trained in a supervised fashion based on paired ground truth training corpora.
One of the main challenges is that paired corpora are often not available.
Recent work on unpaired training of conditional GANs \cite{ZhuPIE2017,YiZTG2017,LiuBK2017,choi2017stargan} removes this requirement.

\paragraph{Generative Models for Humans}

The graphics community has invested significant effort into realistically modeling humans.
Parametric models for individual body parts, such as faces \cite{Blanz99,FLAME:2017}, eyes \cite{Berard14,Wood16}, teeth \cite{Wu16b}, hands \cite{MANO:2017}, as well as for the entire body \cite{Anguelov:2005,SMPL:2015} have been proposed.
Creating a complete photo-real clone of a human based on such models is currently infeasible. 
Recently, generative deep neural networks have been proposed to synthesize 2D imagery of humans.
Approaches that convert synthetic images into photo-realistic imagery have been proposed for eyes \cite{Shrivastava2017}, hands \cite{Mueller2017}, and faces \cite{kim2018DeepVideo}.
The approach of \citet{Ganin2016DeepWarpPI} performs gaze manipulation based on learned image-warping.
In the context of entire human bodies, \citet{Zhu2018} generate novel imagery of humans from different view-points, but cannot control the pose.
In addition, generative models for novel pose synthesis \cite{MaSJSTV2017,SiaroSLS2017,Balakrishnan2018} have been proposed.
The input to such networks are a source image, 2D joint detections, or a stick figure skeleton, and the target pose.
\citet{Balakrishnan2018} train their network end-to-end on pairs of images that have been sampled from action video footage.
Similar to our method, they also use a target body part-based representation as conditional input. However, their body part images are computed by a spatial transformation in 2D, while ours are obtained by rendering the 3D model animated by the target pose. Therefore, our approach yields more plausible body part images as the underlying geometry is correct.
\citet{MaSJSTV2017} combine pose-guided image formation with a refinement network that is trained in an adversarial manner to obtain higher quality.
\citet{SiaroSLS2017} introduce deformable skip connections to better deal with misalignments caused by pose differences.
\citet{Lassner:GP:2017} proposed a 2D generative model for clothing that is conditioned on body shape and pose.
Recently, a VAE for appearance and pose generation \cite{Esser2018} has been proposed that enables training without requiring images of the same identity with varying pose/appearance.
In contrast to previous methods for pose synthesis, our approach employs dense synthetic imagery for conditioning and a novel adversarial loss that dynamically shifts its attention to the regions with the highest photometric residual error. This leads to higher quality results.

\paragraph{Human Performance Capture}

The foundation for realistic video-based characters is reliable and high-quality human performance capture.
The techniques used in professional production are based on expensive photometric stereo \cite{vlasic2009dynamic} or multi-view \cite{matusik2000image,starck2007surface,waschbusch2005scalable} reconstruction setups.
At the heart of most approaches is a skeletal motion prior \cite{gall2009motion,vlasic2008articulated,liu2011markerless}, since it allows to reduce the number of unknown pose parameters to a minimum.
A key for reliable tracking and automatic initialization of performance capture methods is the incorporation of 2D \cite{pishchulin2016deepcut,wei2016convolutional} and 3D \cite{zhou2016deep,mehta2016monocular,pavlakos2016coarse} joint detections of a pose estimation network into the alignment objective.
Hybrid methods \cite{elhayek2015efficient,rosales2006combining,VNect_SIGGRAPH2017} that rely on 2D as well as 3D detections combine both of these constraints for higher quality.
High-quality reconstruction of human performances from two or more cameras is enabled by model-based approaches \cite{cagniart2010free,de2008performance,wu2013onset}.
Currently, the approaches that obtain the highest accuracy are multi-view depth-based systems, e.g., \cite{collet2015high,Dou:2016,Dou:2017,wang2016capturing}.
Driven by the demand of VR and AR applications, the development of lightweight \cite{Zhang2014,bogo2015detailed,Helten:2013,yu2017bodyfusion,bogo2016smpl} solutions is an active area of research and recently even monocular human performance capture has been demonstrated \cite{MonoPerfCap_SIGGRAPH2018}.

\section{Method}
\label{sec:method}

\begin{figure*}
	\includegraphics[width=\linewidth]{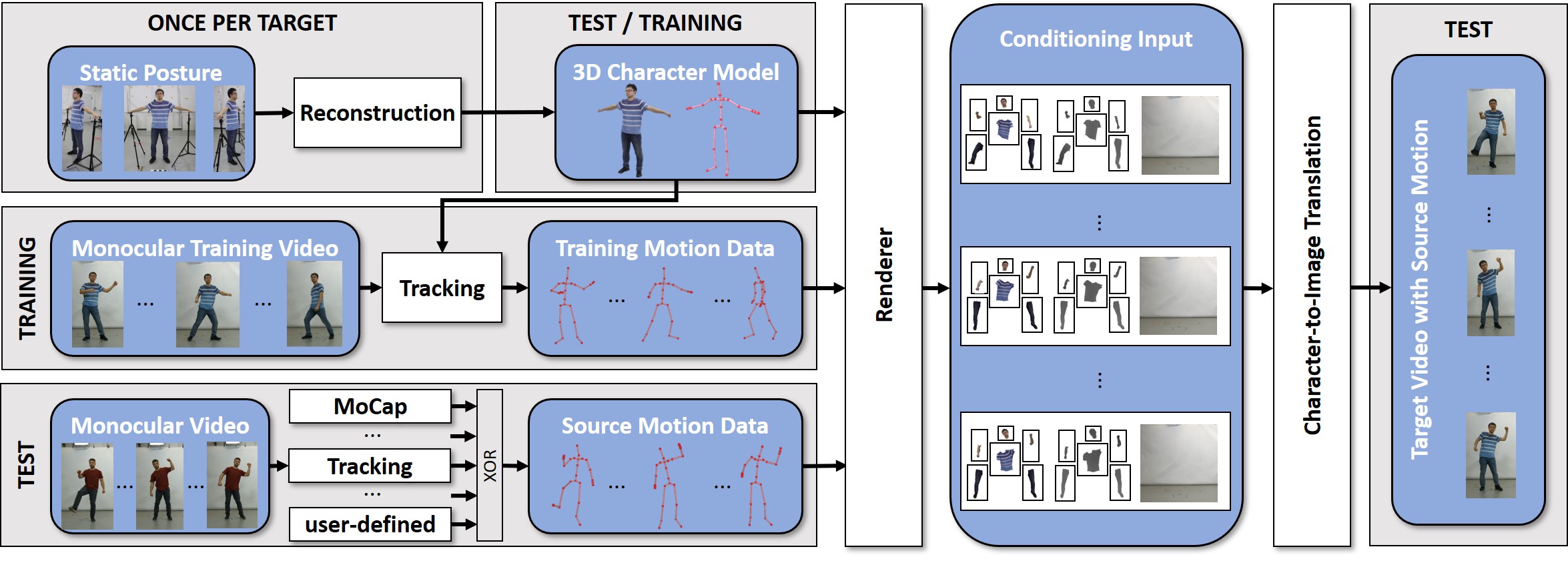}
	\vspace{-0.8cm}
    \caption{
    	Overview of our proposed approach for neural reenactment of human performances. Using a 3D character model of the target person, which is reconstructed from static posture images, we obtain the training motion data from the monocular training video based on an extension of the method of \cite{VNect_SIGGRAPH2017}. The motion data is then used to steer the 3D character model, from which we render color and depth body part images, which are fed into our Character-to-Image translation network in order to produce video-realistic output frames. At test time, the source motion data is used to render the conditional input (with the target 3D character model) in order to reenact the target human character. Note that the source motion can either come from a tracked monocular video, or from any other motion source (e.g., user-defined motion, or MoCap data).} 
    \label{fig:overview}
\end{figure*}

In this section we describe the technical details of our approach, which is outlined in Fig.~\ref{fig:overview}.
The main idea is to train a neural network that converts simple synthetic images of human body parts to an image of a human character that exhibits natural image characteristics such that the person appears (close to) photo-realistic.
The main motivation for our approach is as follows: 
Photo-realistic rendering based on the conventional graphics pipeline requires expensive high-quality 3D human character modeling and the simulation of complex global light transport.
In contrast, it is relatively simple and cheap to construct medium-quality synthetic human imagery based on commodity 3D reconstructions and direct illumination models.
In order to compensate for the lack of photo-realism in such medium-quality images, we propose to train a generative deep neural network to bridge this gap, such that the person looks more realistic.
In the following, we first describe the acquisition of suitable training data, followed by an in-depth explanation of the architecture of our Character-to-Image translation network.

\subsection{Acquisition of the Training Corpus}
\label{sec:trainingData}

In this section we describe how we acquire our training corpus.
Our training corpus consists of pairs of rendered conditioning input images and the original image frames of a monocular training video (cf.~Fig.~\ref{fig:overview}). The conditioning images that are used as input for the network comprise individual depth and color renderings of six human body parts, i.e., \emph{head, torso, left arm, right arm, left leg} and \emph{right leg}, and an empty background image. In the following we describe how these images are obtained.
\paragraph{Data Acquisition.} We capture our raw data using a Blackmagic video camera. For each actor, we record a motion sequence of approximately 8 minutes (about 12k frames). Similar to most learning based methods, our training data should resemble the distribution of real-world observations. Therefore, for each subject, we collect the training video such that it covers a typical range of general motions.
\paragraph{3D Character Model.} Our method relies on a textured 3D template mesh of the target subject, which we obtain by capturing (around) 100 images of the target person in a static pose from different viewpoints, and then reconstructing a textured mesh using a state-of-the-art photogrammetry software\footnote{Agisoft Photoscan, http://www.agisoft.com/}. The images need to be captured in such a way that a complete 3D reconstruction can be obtained, which is achieved by using a hand-held camera and walking around the subject.
Afterwards, the template is rigged with a parameterized human skeleton model. More details on the template reconstruction can be found in \cite{MonoPerfCap_SIGGRAPH2018}.

\paragraph{Conditioning Input Images.} In order to obtain the conditioning input images, we track the skeleton motion of the person in the training video using the skeletal pose tracking method of \cite{VNect_SIGGRAPH2017}.
We extended this approach by a dense silhouette alignment constraint and a sparse feature alignment term based on a set of detected facial landmarks \cite{SaragLC2011}.
Both additional terms lead to a better overlap of the model to the real-world data, thus simplifying the image-to-image translation task and leading to higher quality results.
The output of the method is a sequence of deformed meshes, all of them sharing the same topology (cf. \emph{Training/Source Motion Data} in~Fig.~\ref{fig:overview}).
We apply temporal smoothing to the trajectories of all vertices based on a Gaussian filter (with a standard deviation of~$1$ frame).
We then generate three different types of conditioning images by rendering synthetic imagery of the mesh sequence using the 3D character model,
Specifically, we render (i) the textured mesh to obtain the color image $\mathcal{I}$, (ii) the depth image $\mathcal{D}$, and (iii) the binary semantic body part masks $\{\mathcal{M}_p\;|\;p\in \{1,...,6\}\}$.
To render the binary semantic body part masks, we manually labeled the $6$ body parts \emph{head, torso, left arm, right arm, left leg, right leg} on the template mesh, and then generated a binary mask for each individual body part.
Based on these masks, we extract the part-based color images $\{\mathcal{I}_p = \mathcal{I} \odot \mathcal{M}_p \;|\;p\in \{1,...,6\}\}$ and the part-based depth images $\{\mathcal{D}_p = \mathcal{D} \odot \mathcal{M}_p \;|\;p\in \{1,...,6\}\}$, where $\odot$ denotes the Hadamard product.
Finally, we generate the conditioning input images by concatenating the $6$ part-wise color images $\{\mathcal{I}_p\}$ and depth images $\{\mathcal{D}_p\}$, as well as the empty background image $\mathcal{B}$, along the channel axis  (cf. \emph{Conditioning Input} in~Fig.~\ref{fig:overview}), resulting in the input $\mathbf{X}$. More details are described in the supplementary material.


\subsection{Character-to-Image Translation}
\label{sec:translation}
\begin{figure*}[t!]
	\includegraphics[width=\linewidth]{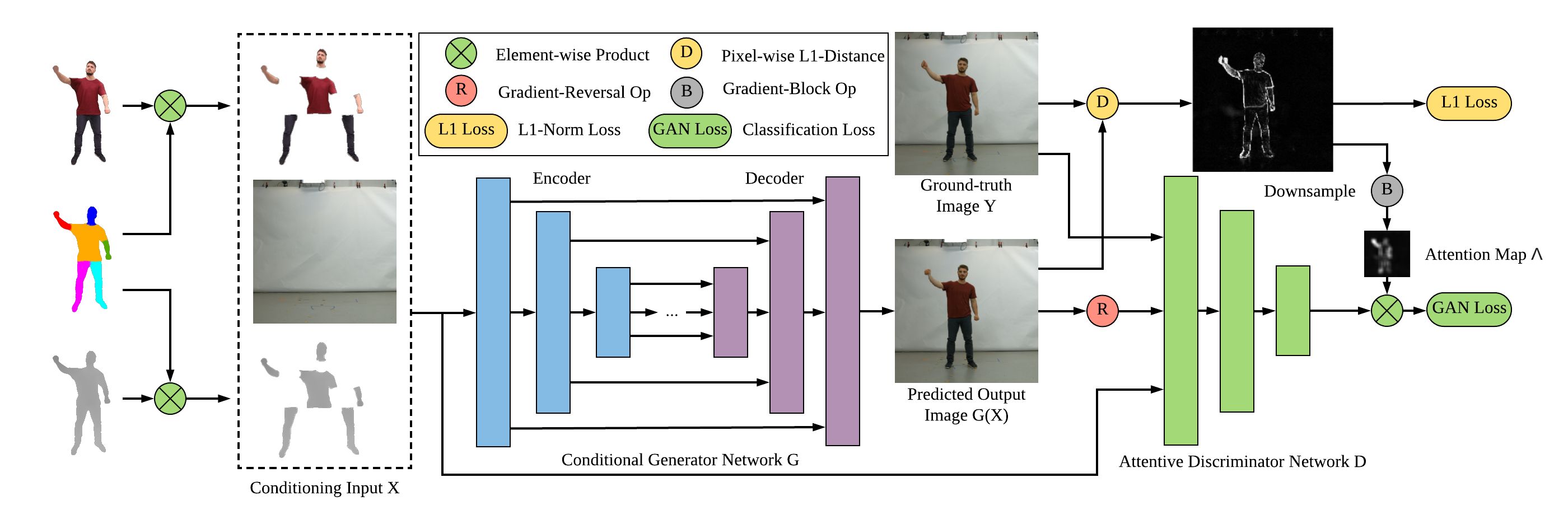}
	\caption{
		An illustration of the architecture of our Character-to-Image translation network.
        Both the encoder and the decoder of the generator have 8 layers with skip-connections between the same resolution levels.
        We use the gradient-reversal operation to reverse the sign of the gradients flowing into the generator and the gradient-block operation to stop backpropagation from the downsampled error map to avoid a cheating path.
	}
	\label{fig:architecture}
\end{figure*}

In this section we describe our \emph{Character-to-Image} translation network (Fig.~\ref{fig:architecture}) in detail.

\paragraph{Motion transfer}
In order to synthesize a video of the target person mimicking the motion of the source person, we transfer the per-frame skeletal pose parameters $\mathcal{P}_{\text{src}}$ (comprising the global location, global orientation and individual joint angles) from the source $\mathcal{S}_{\text{src}}$ to the target $\mathcal{S}_{\text{trgt}}$, where $\mathcal{S}_*$  denotes the skeleton model (comprising the skeleton topology and the bone lengths)~\cite{MonoPerfCap_SIGGRAPH2018}. 
Afterwards, we animate the virtual character of the target person with the transferred poses and finally render the target character to generate the conditioning images.
We point out that we do not directly apply the source's skeletal pose parameters $\mathcal{P}_{\text{src}}$ to the target's skeleton $\mathcal{S}_{\text{trgt}}$ due to two issues: on the one hand, this would require that both skeletons have exactly the same structure, which may be overly restrictive in practical applications. On the other hand, and more importantly, differences in the rigging of the skeleton would lead to incorrect poses.
To address these issues, we estimate the optimal pose $\mathcal{P}^*$ of the target person for each frame by solving the following inverse kinematics (IK) problem, which encourages that (corresponding) keypoints on both skeletons, including the joints and facial landmarks, match in 3D:
s\begin{equation}
	\mathcal{P}^* = \argmin_{\mathcal{P}} \sum_{k\in K} \big|\big| \mathbf{J}_k \big(\mathcal{S}_{\text{src}}, \mathcal{P}_{\text{src}}\big) - \mathbf{J}_k\big( \mathbf{\Phi}(\mathcal{S}_\text{trgt}, \mathcal{S}_{\text{src}}), \mathcal{P}\big) \big|\big|_2^2
\end{equation}
Here $K$ denotes the number of keypoints on the skeleton,
$\mathbf{J}_k(\mathcal{S}_*, \mathcal{P}_*)$ is a function that computes the 3D position of the $k$-th keypoint given a skeleton $\mathcal{S}_*$ and a pose $\mathcal{P}_*$, and the function $\mathbf{\Phi}(\mathcal{S}_\text{trgt}, \mathcal{S}_\text{src})$ returns the skeleton $\mathcal{S}_\text{trgt}$ after  each individual bone length of $\mathcal{S}_\text{trgt}$ has been rescaled to match $\mathcal{S}_\text{src}$.
To ensure that $\mathcal{S}_\text{trgt}$ is globally at a similar position as in the training corpus, we further translate $\mathcal{S}_\text{trgt}$ by a constant offset calculated with the root position of the skeleton in the test sequence and training sequence.
Note that this IK step enables motion transfer between skeletons with different structures, and thus allows us to use motion data from arbitrary sources, such as artist designed motions or MoCap data, to drive our target character.

\paragraph{Network Architecture}

Our Character-to-Image translation network (Fig.~\ref{fig:architecture}) consists of two competing networks, a conditional generator network $\mathbf{G}$ and an attentive discriminator network $\mathbf{D}$ based on the attention map $\boldsymbol{\Lambda}$. 

The purpose of the generator network $\mathbf{G}$ (cf.~\emph{Generator Network} in~Fig.~\ref{fig:architecture}) is to translate the input $\mathbf{X}$, which comprises synthetic color and depth renderings of $6$ human body parts and the background image, as described in Sec.~\ref{sec:trainingData}, to a photo-realistic image $\mathbf{G}(\mathbf{X})$ of the full character. 
Our generator network internally consists of an encoder (cf.~\emph{Encoder} in~Fig.~\ref{fig:architecture})  to compress the input into a low-dimensional representation, and a decoder (cf.~\emph{Decoder} in~Fig.~\ref{fig:architecture}) to synthesize the photo-realistic image conditioned on the input character renderings.
Each encoder layer comprises a $4 \times 4$ convolution with stride 2, batch normalization and a leaky Rectified Linear Unit (ReLU) activation.
The decoder reverses the downsampling due to the encoding process with a symmetric $4 \times 4$ deconvolution operator with stride 2, which is fed into batch normalization, dropout and ReLU layers.
In order to ensure that the final output of the network, i.e., the generated image (cf.~\emph{Predicted Output Image in~Fig.~\ref{fig:architecture})}, is normalized, we apply a hyperbolic tangent activation at the last layer. 
In addition, skip connections \cite{RonneFB2015} and a cascaded refinement strategy \cite{ChenK2017} are used in the generator network to propagate high-frequency details through the generator network.
Both the input and output images are represented in a normalized color space, i.e., $[-1,-1,-1]$ and $[1,1,1]$ for black and white respectively.

The input to our attentive discriminator network $\mathbf{D}$ is the conditioning input $\mathbf{X}$, and either the predicted output image $\mathbf{G}(\mathbf{X})$ or the ground-truth image $\mathbf{Y}$ (cf.~Fig.~\ref{fig:architecture}). 
The employed discriminator is motivated by the PatchGAN classifier \cite{IsolaZZE2017}, which we extend to incorporate an attention map to reweigh the classification loss.  For more details, please refer to Fig. \ref{fig:architecture}.

\paragraph{Objective Function}
In order to achieve high-fidelity character-to-image translation, we base the objective on the expected value of our attentive conditional GAN loss $\mathcal{L}$ and on the $\ell_1$-norm loss $\mathcal{L}_{\ell_1}$:
\begin{equation}
\label{eq.loss}
\min_{\mathbf{G}}{\max_{\mathbf{D}}{ \mathbb{E}_{\mathbf{X},\mathbf{Y}}\big[ \mathcal{L}(\mathbf{G}, \mathbf{D}) + \lambda \mathcal{L}_{\ell_1}(\mathbf{G})}} \big]\text{.}
\end{equation}

The $\ell_1$-distance of the synthesized image $\mathbf{G}(\mathbf{X})$ from the ground-truth image $\mathbf{Y}$ is introduced so that the synthesized output is sufficiently sharp while remaining close to the ground truth:
\begin{equation}
\mathcal{L}_{\ell_1}(\mathbf{G}) = 
 \left\lVert \mathbf{Y} - \mathbf{G}(\mathbf{X}) \right\rVert_1\text{.}
\end{equation}

\paragraph{Attentive Discriminator}
One of the technical novelties of our approach is an attentive discriminator to guide the translation process.
As the attention map is used to reweight the discriminator loss, it is downsampled to the resolution of the discriminator loss map.
Similar to the vanilla PatchGAN classifier \cite{IsolaZZE2017}, the discriminator $\mathbf{D}$ predicts a $W_{D}\times H_{D}$ map, in our case 30 $\times$ 30,
 where each value represents the probability of  the receptive patch for being \textit{real}, i.e., value $1$ means that the discriminator decides that the patch is \emph{real}, and the value $0$ means it is \emph{fake}.
However, in contrast to the PatchGAN approach that treats all patches equally, we introduce the attention map $\boldsymbol{\Lambda}$, which has the same spatial resolution as the output of $\mathbf{D}$, i.e., $W_{D}\times H_{D}$. Its purpose is to shift the focus of areas that $\mathbf{D}$ relies on depending on some measure of importance, which will be discussed below.
For the GAN loss, the discriminator $\mathbf{D}$ is trained to classify between \textit{real} and \textit{fake} images given the synthetic character input $\mathbf{X}$, while the generator $\mathbf{G}$ tries to fool the discriminator network by sampling images from the distribution of \textit{real} examples:
\begin{equation}
\label{eq.ganAttentive}
\mathcal{L}(\mathbf{G},\mathbf{D}) = 
\sum \left(
\frac{\boldsymbol{\Lambda}}{\|\boldsymbol{\Lambda}\|} \odot \left( \log \mathbf{D}(\mathbf{X},\mathbf{Y}) + \log\big(\mathbf{1} {-} \mathbf{D}(\mathbf{X},\mathbf{G}(\mathbf{X})\big)\right) \right).
\end{equation}
Here, $\log$ is element-wise, by $\|\boldsymbol{\Lambda}\|$ we denote a scalar normalization factor that sums over all entries of $\boldsymbol{\Lambda}$, and the outer sum $\sum(\cdot)$ sums over all $W_D \times H_D$ elements due to the matrix-valued discriminator. 

We have found that a good option for choosing the attention map $\boldsymbol{\Lambda}$ is to use the model's per-pixel $\ell_1$-norm loss after downsampling it to the resolution $W_{D}\times H_{D}$.
The idea behind this is to help $\mathbf{G}$ and $\mathbf{D}$ focus on parts where $\mathbf{G}$ still produces low quality results.
For instance, $\mathbf{G}$ quickly learns to generate background (since, up to shadows or interactions, it is almost fixed throughout training and can be captured easily through skip-connections.), which leads to a very small $\ell_1$-norm loss, and thus fools $\mathbf{D}$.
However, there is no explicit mechanism to stop learning $\mathbf{G}$ in these regions, as $\mathbf{D}$ still tries to classify \textit{real} from \textit{fake}.
These ``useless'' learning signals distract the gradients that are fed to $\mathbf{G}$ and even affect the learning of other important parts of the image.
In this case, the $\ell_1$-norm loss is a good guidance for GAN training. 

\paragraph{Training}
In order to train the proposed character-to-image translation network, we use approximately $12{,}000$ training pairs, each of which consists of the original monocular video frame $\mathbf{Y}$ as well as the stack of conditioning images $\mathbf{X}$, as described in Sec.~\ref{sec:trainingData}.
For training, we set a hyper-parameter of $\lambda \!=\! 100$ for the loss function (Eq.~\ref{eq.loss}), and use the Adam optimizer ($lr=0.0002$, $\beta_1=0.5$, $\beta_2=0.99$) from which we run for a total of $70{,}000$ steps with a batch size of $10$. The number of layers in the generator was empirically determined. We implemented our model in TensorFlow \citep{AbadiABBCCCDDDGGHIIJJKKLMMMMOSSSSTTVVVVWWWYZ2015}.

\section{Experiments}
\label{sec:experiments}

\begin{figure*}[t!]
	\rotatebox[origin=l]{90}{~~\textbf{Output} (synth.) \qquad$ \textbf{Rendered mesh} \quad$ \textbf{Driving motion}}
	\includegraphics[scale=.3]{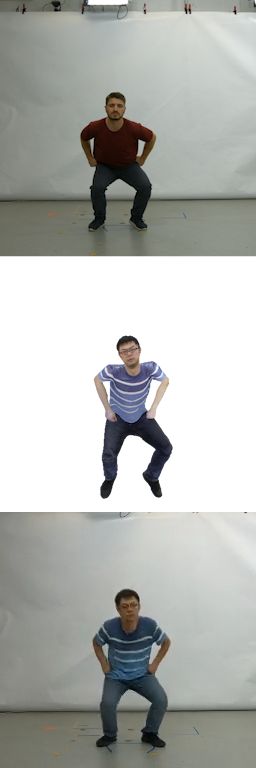}	\hspace{1.2mm}
	\includegraphics[scale=.3]{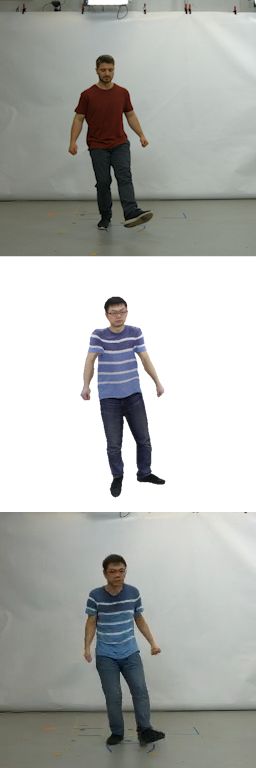} \hspace{1.2mm}	
	\includegraphics[scale=.3]{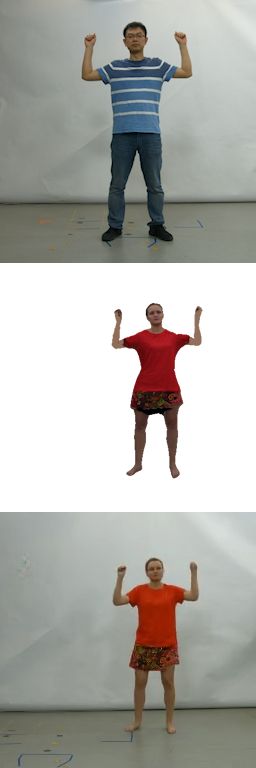}	\hspace{1.2mm}
    \includegraphics[scale=.3]{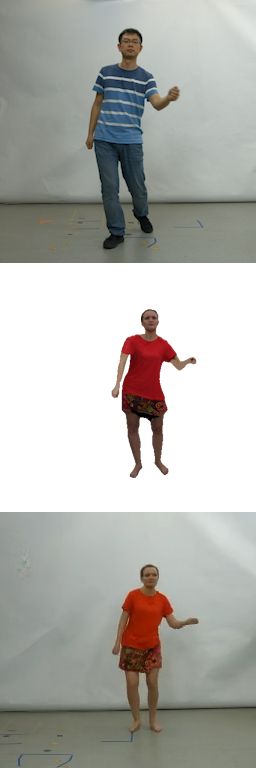}	\hspace{1.2mm}
	\includegraphics[scale=.3]{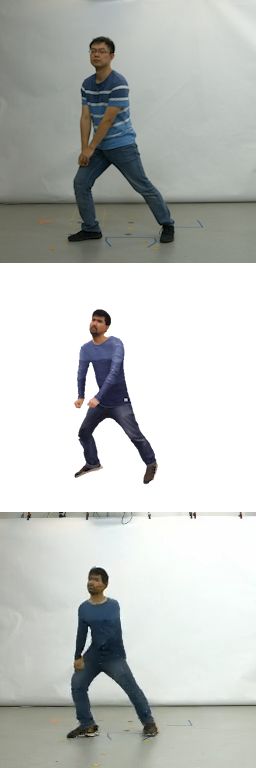}	\hspace{1.2mm}
    \includegraphics[scale=.3]{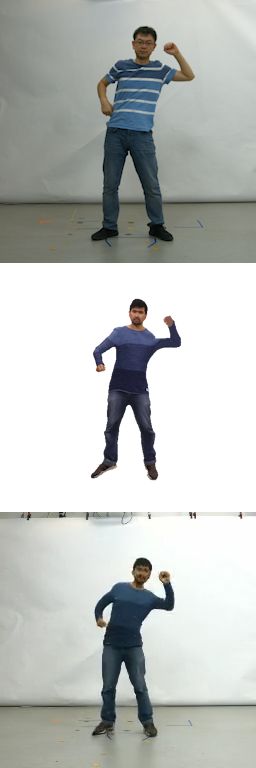}	
    \caption{
		Qualitative reenactment results. The images in the first row (``Driving motion'') are the input to our approach and define the motion, from which the conditioning images are generated.
		The images in the middle row (''Rendered Mesh") show the 3D mesh of the target person driven by the source motion.
		The images in the bottom row (``Output'') are synthesized by our approach.
		Our method is able to faithfully transfer the input motion onto the target character for a wide range of human body performances.
		For full results we refer to the supplementary video.
		}
	
	\label{fig:qualitative-allinputs}
  
\end{figure*}
\begin{figure*}[t]
	\rotatebox[origin=l]{90}{\textbf{Output} (synth.) $\qquad$ \textbf{Driving motion}} 
	\includegraphics[scale=.3]{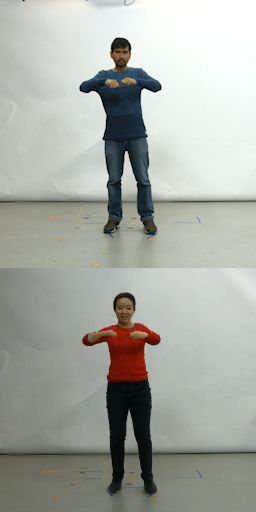}	\hspace{1.2mm}
	\includegraphics[scale=.3]{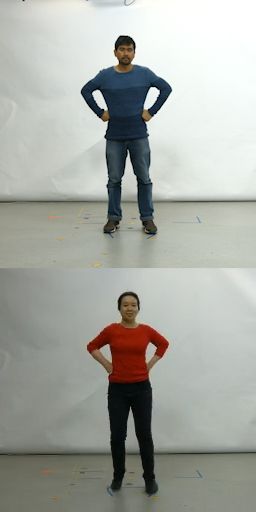}	\hspace{1.2mm}
	\includegraphics[scale=.3]{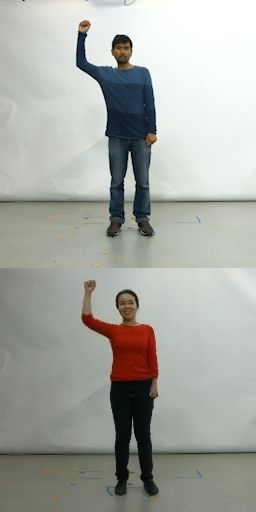}	\hspace{1.2mm}
	\includegraphics[scale=.3]{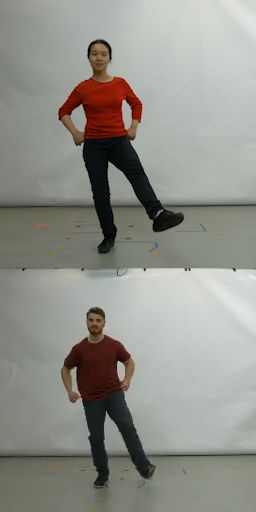}	\hspace{1.2mm}
	\includegraphics[scale=.3]{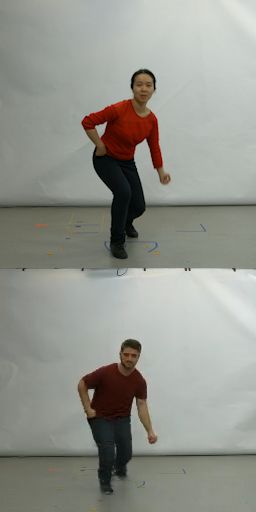}	\hspace{1.2mm}
	\includegraphics[scale=.3]{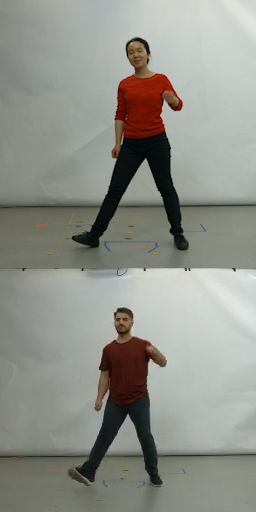}	
    \\
	\vspace{2mm}
    \rotatebox[origin=l]{90}{\textbf{Output} (synth.) $\qquad$ \textbf{Driving motion}}
        \includegraphics[scale=.3]{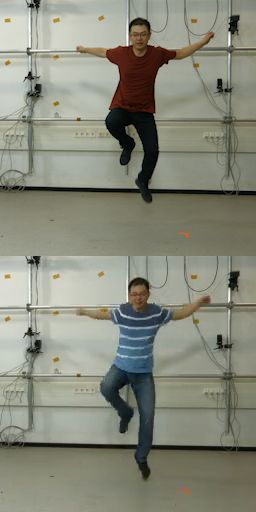}	\hspace{1.2mm}
	\includegraphics[scale=.3]{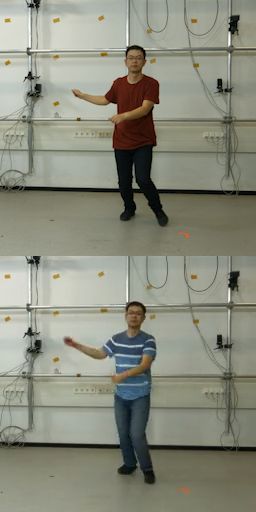}	\hspace{1.2mm}
    \includegraphics[scale=.3]{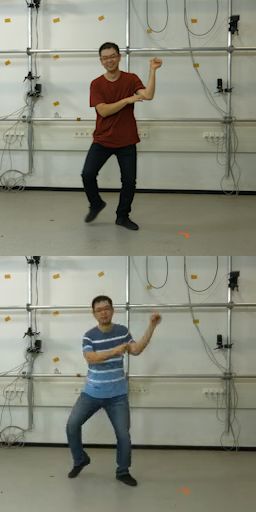}	
    	\hspace{1.2mm}
	\includegraphics[scale=.3]{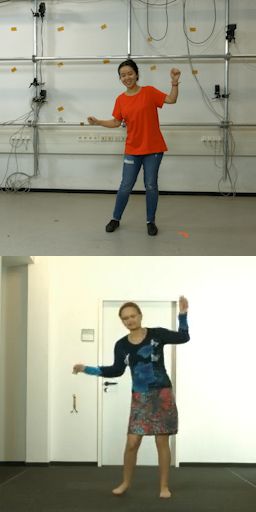}	\hspace{1.2mm}
	\includegraphics[scale=.3]{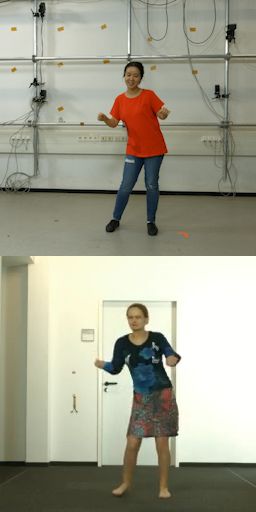} \hspace{1.2mm}	
	\includegraphics[scale=.3]{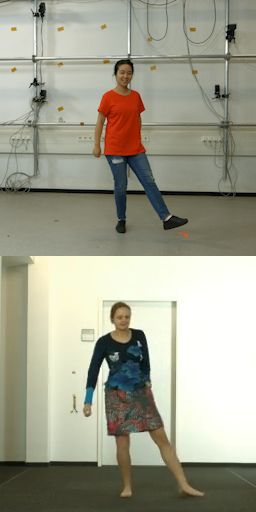}
	\caption{ 
		Qualitative results for the reenactment of human characters.
		The images in the ``Driving motion'' row are used to track the motion, from which the conditioning images are generated.
		The images in the ``Output'' row are synthesized by our approach.
		Our method is able to faithfully transfer the input motions onto the target character for a wide range of human body performances.
		For full results we refer to the supplementary video.
		}
		
	\label{fig:qualitative}
\end{figure*}

\newcommand{\scalecomparison}{.28}
\newcommand{\hspacecomparison}{1.5mm}
\begin{figure*}
	\rotatebox[origin=l]{90}{\cite{Esser2018} \hspace{.2cm} \cite{MaSGVSF2017} \hspace{1cm} \textbf{Ours} \hspace{1.5cm} \textbf{Input}} 
	\includegraphics[scale=\scalecomparison]{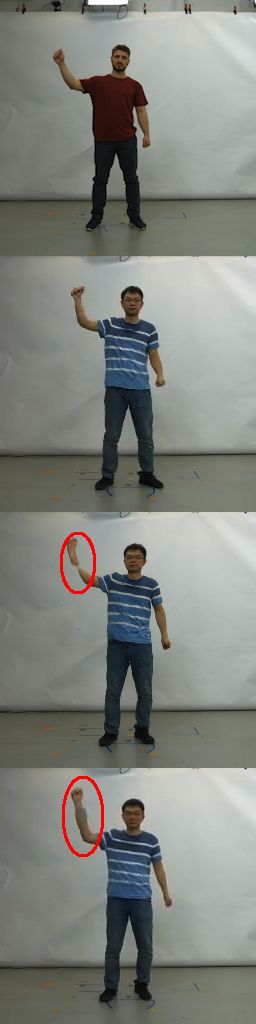} \hspace{\hspacecomparison}
	\includegraphics[scale=\scalecomparison]{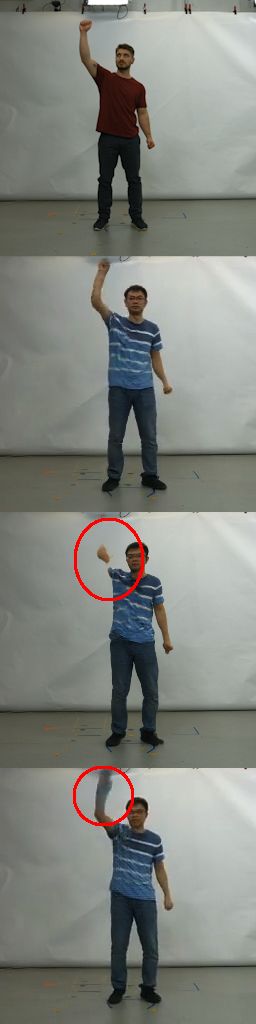}	\hspace{\hspacecomparison}
	\includegraphics[scale=\scalecomparison]{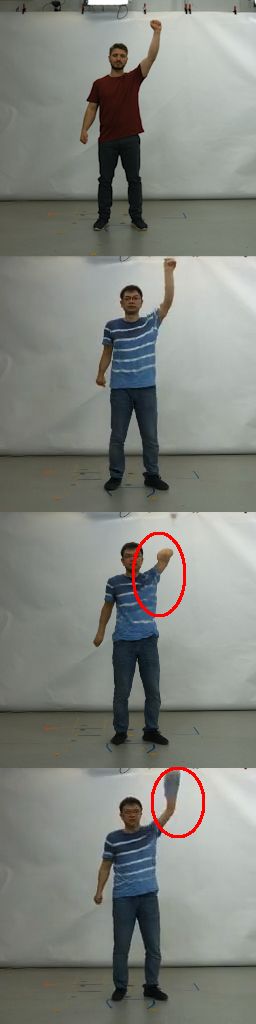}	\hspace{\hspacecomparison}
	\includegraphics[scale=\scalecomparison]{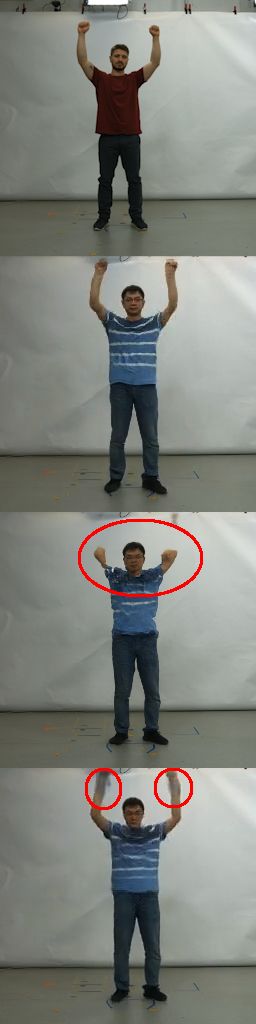}	\hspace{\hspacecomparison}
	\includegraphics[scale=\scalecomparison]{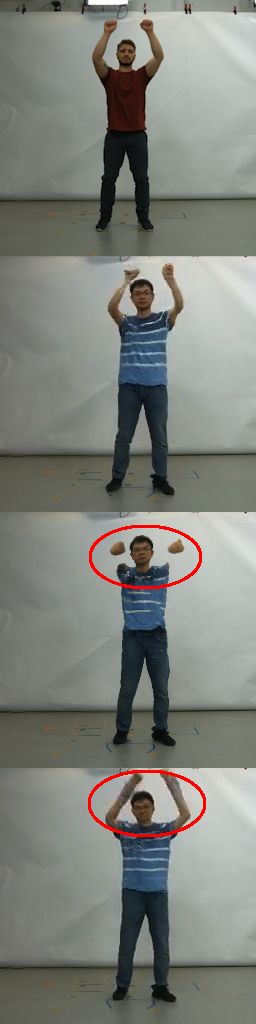}	\hspace{\hspacecomparison}
 	\includegraphics[scale=0.374]{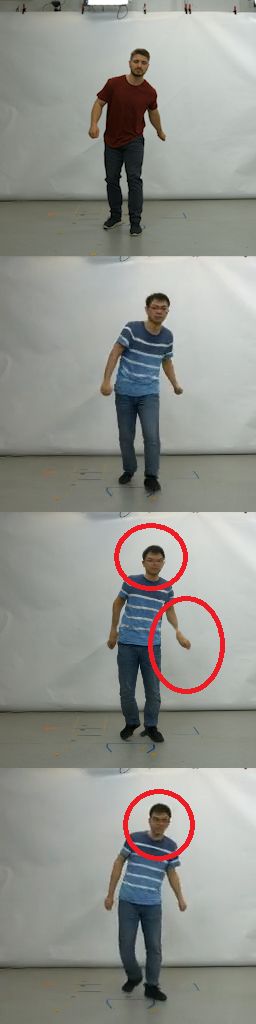}	
	\\
	\vspace{5mm}
		\rotatebox[origin=l]{90}{\cite{Esser2018} \hspace{.2cm} \cite{MaSGVSF2017} \hspace{1cm} \textbf{Ours} \hspace{1.5cm} \textbf{Input}}
	\includegraphics[scale=\scalecomparison]{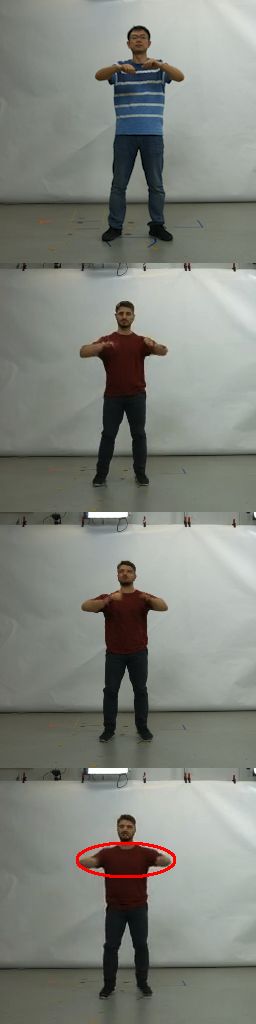}	\hspace{\hspacecomparison}
	\includegraphics[scale=\scalecomparison]{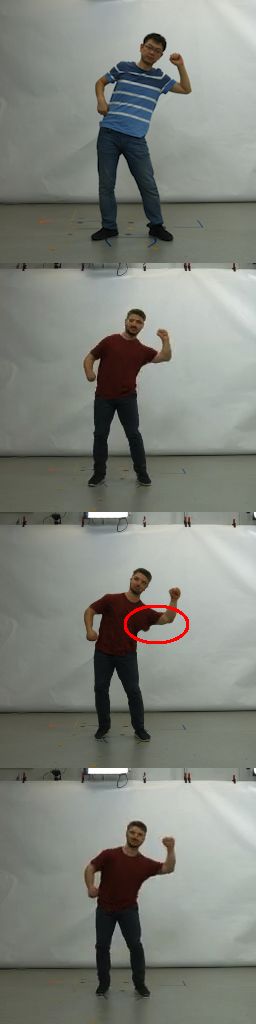}	\hspace{\hspacecomparison}
	\includegraphics[scale=\scalecomparison]{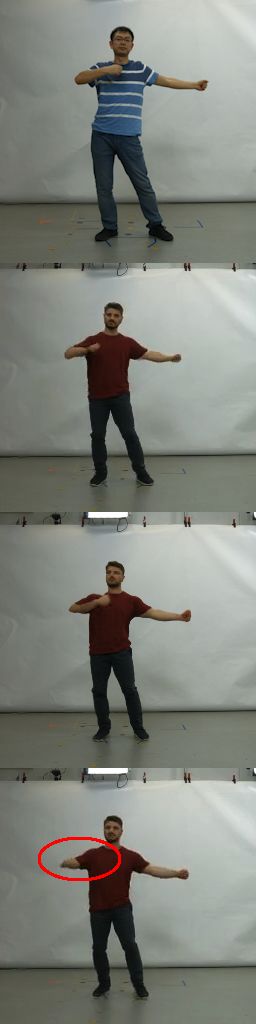}	\hspace{\hspacecomparison}
	\includegraphics[scale=\scalecomparison]{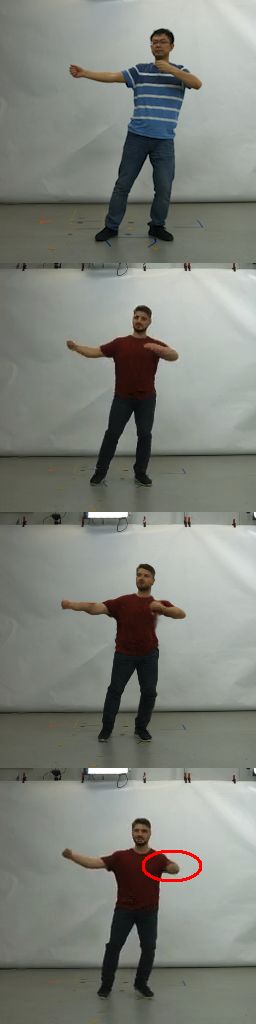}	\hspace{\hspacecomparison}
	\includegraphics[scale=\scalecomparison]{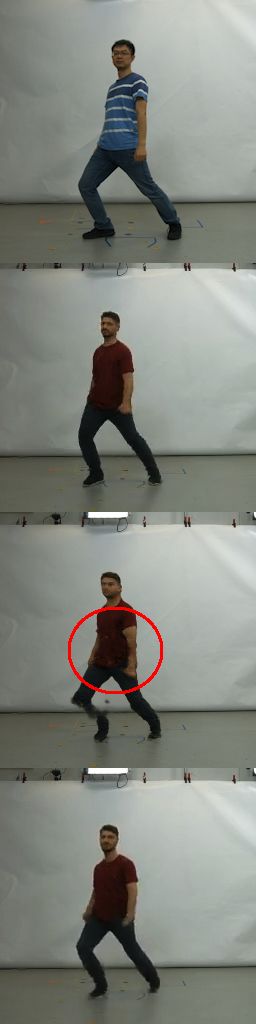}	\hspace{\hspacecomparison}
	\includegraphics[scale=\scalecomparison]{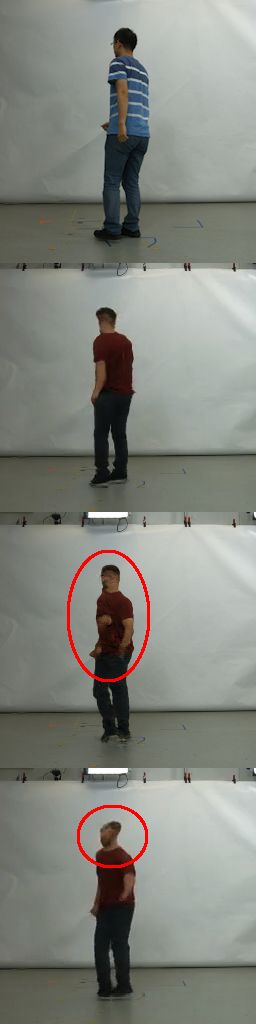}	
	\vspace{-2mm}
	\caption{
		Qualitative re-targeting comparison of our approach to the state-of-the-art human image synthesis methods of \citet{MaSGVSF2017} and \citet{Esser2018}. 
The encircled areas show regions of the body where the other methods clearly fail (missing parts of arms, wrongly shaped body parts etc.) in comparison to our approach. 
        For the full results we refer to the supplementary video.
	}
	\label{fig:comparison-sota}
\end{figure*}

In order to evaluate our approach, we captured training sequences of 5 subjects performing various motions.
Our training corpus consists of approximately $12{,}000$ frames per person.
We further recorded 5 separate source video sequences for motion retargeting.
We perform pose tracking using~\citet{VNect_SIGGRAPH2017} to obtain the driving skeletal motion.
Training takes about 12h for each subject on a resolution of $256 \times 256$ pixels using a single NVIDIA Tesla V100.
Template rendering takes less than 1ms/frame.
A forward pass of our network takes about 68ms per frame.

\begin{figure}[t]
\includegraphics[width=0.93\columnwidth]{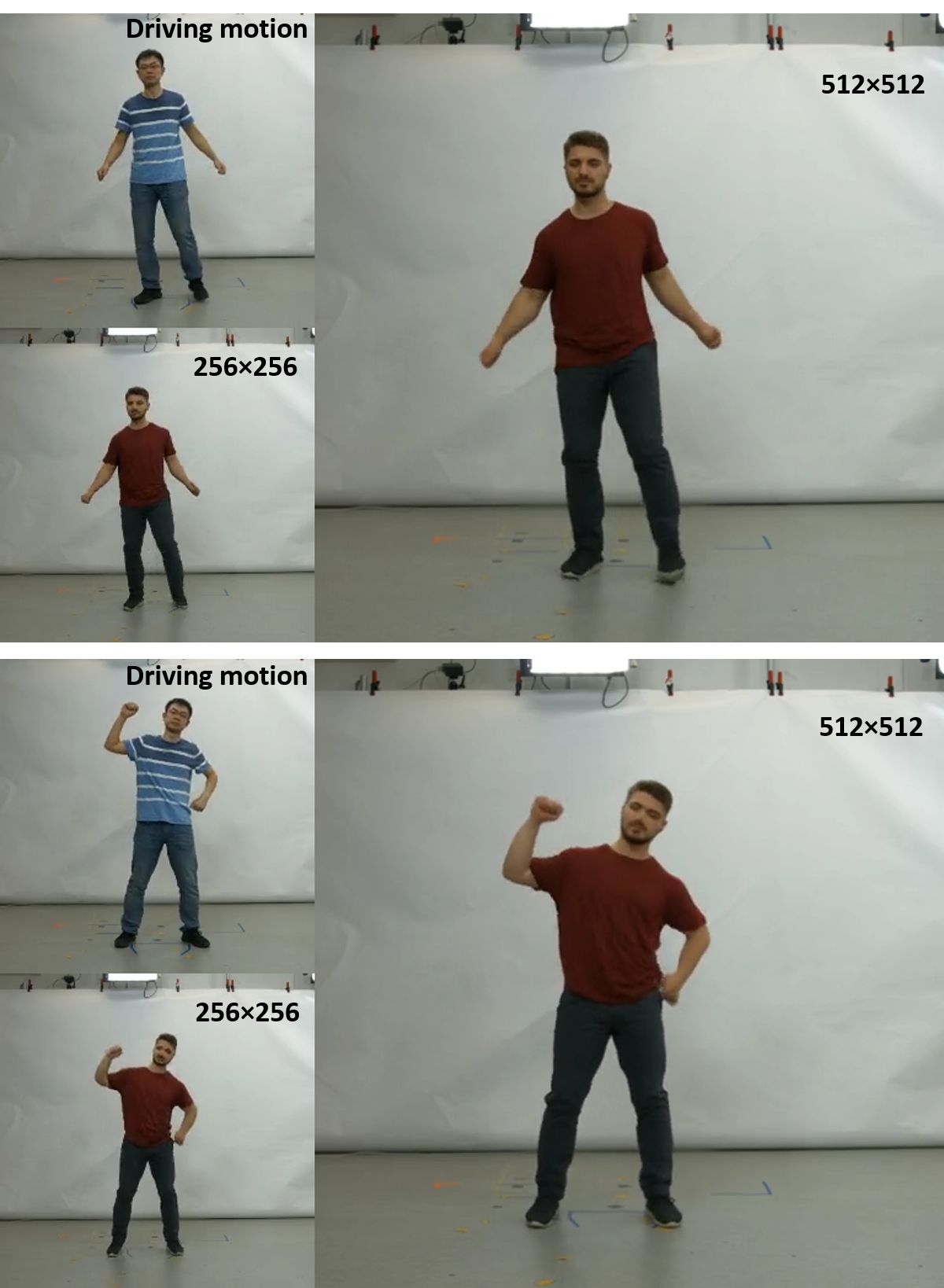}	
		\vspace{-0.3cm}
	\caption{
		Our approach can also synthesize high resolution images ($512 \times 512$ pixels). This further improves the sharpness of the results but comes at the cost of an increased training time of the network.
	}
	\label{fig:highres}
\end{figure}

The results presented in the following have been generated with a resolution of $256 \times 256$ pixels.
Please note, that our approach is also able to generate higher resolution results of 512 $\times$ 512 pixels.
Fig.~\ref{fig:highres} shows a few such examples, which took ~24h to train.
This further improves the sharpness of the results.
Our dataset and code 
will be made publicly available.

In the following, we evaluate our method qualitatively and quantitatively.
We provide a comparison to two state-of-the-art human image synthesis methods trained on our dataset.
We also perform an ablation study to analyze the importance of each component of our proposed approach.

\subsection{Qualitative Results}
Figs.~\ref{fig:qualitative-allinputs} and ~\ref{fig:qualitative} show example reenactment results.
%
%
We can see that our method synthesizes faithful imagery of human performances, in which the target characters precisely reenact the motion performed by the source subject.
Our final results add a significant level of realism compared to the rendered character mesh from which the conditioning input is computed.
Our method generalizes to different types of motions, such as waving, boxing, kicking, rotating and many gymnastic activities.
Note that our method generates sharp images with a large amount of fine-scale texture detail.
The facial features and the textures on the clothing are both well-preserved.
Also note that our results accurately resemble the illumination condition in the real scene.
Even the shading due to wrinkles on the garments and the shadows cast by the person onto the ground and wall are consistently synthesized.

\begin{figure}[t!]
\includegraphics[width=0.8\columnwidth]{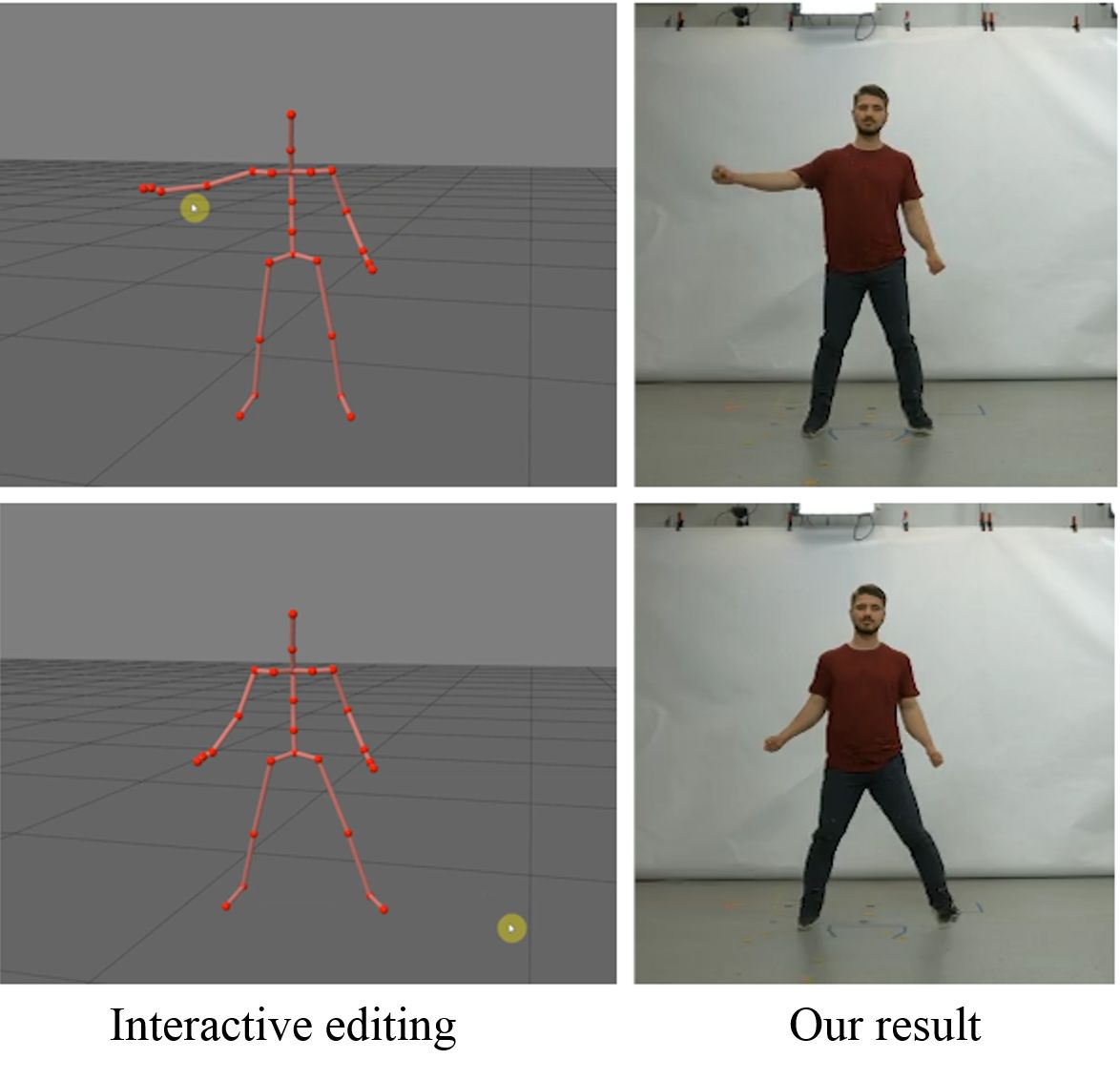}	
		\vspace{-0.3cm}
	\caption{
		Interactive editing results. Since a forward pass of our network only takes about 68ms, our approach allows for interactive user-controlled pose editing in video.
	}
	\label{fig:interactive editing}
\end{figure}

Since a forward pass of our character-to-image translation network requires only 68\,ms, it can also be used to generate new unseen poses based on interactive user control.
Fig.~\ref{fig:interactive editing} shows a few examples of a live editing session, where a user interactively controls the skeleton pose parameters of a real-world character using handle-based inverse kinematics.
Please refer to the accompanying video for the complete video result.

Next, we compare our approach to the state-of-the-art human body image synthesis methods of~\citet{MaSGVSF2017} and~\citet{Esser2018}, which we also trained on our dataset.
For a fair comparison, we trained one person-specific network per subject for both the method of~\citet{MaSGVSF2017} and of~\citet{Esser2018}, as done in our approach.
Note that their methods take 2D joint detections as conditioning input.
However, using the 2D joint detection of the source subject could make their methods produce inaccurate results, since during training the networks only see the skeleton of the target subject, which may have a different spatial extent than the source skeleton.
Hence, to obtain a fair comparison, we use our transferred motion applied to the target subject (see Sec.~\ref{sec:translation}) to generate the 2D joint positions.
A qualitative comparison is shown in Fig.~\ref{fig:comparison-sota}.
We can see that the results of~\citet{MaSGVSF2017} and~\citet{Esser2018} exhibit more artifacts than the outputs produced by our approach.
In particular, both \citet{MaSGVSF2017} and~\citet{Esser2018} have difficulties in faithfully reproducing strongly articulated areas, such as the arms, as well highly textured regions, such as the face. 
In contrast, our method results in shaper reconstructions, preserves more details in highly textured regions such as the face, and leads to fewer missing parts in strongly articulated areas, such as the arms.

\subsection{User Study}
In order to evaluate the user perception of the motion reenactment videos synthesized by our approach, we have conducted a user study that compares our results with the results obtained by \citet{MaSGVSF2017} and~\citet{Esser2018}.
To this end, we present pairs of short video clips, approximately of length between $4$ and $25$ seconds, to a total of $25$ users, recruited mainly from Asia and Europe.
We used a total of $14$ sequence pairs, where for each pair exactly one sequence was produced by our method, whereas the other sequence in the pair was produced either by \citet{MaSGVSF2017} or by~\citet{Esser2018}, each of them being used $7$ times.
The users were asked to select for each pair the sequence which appears more realistic. 
In total, in $84\%$ of the $25{\cdot}14 = 350$ ratings our method was preferred, whereas in $16\%$ of the ratings one of the other methods was preferred.

\subsection{Ablation Study}

\begin{figure*}[t!]
	\includegraphics[width=2\columnwidth]
	{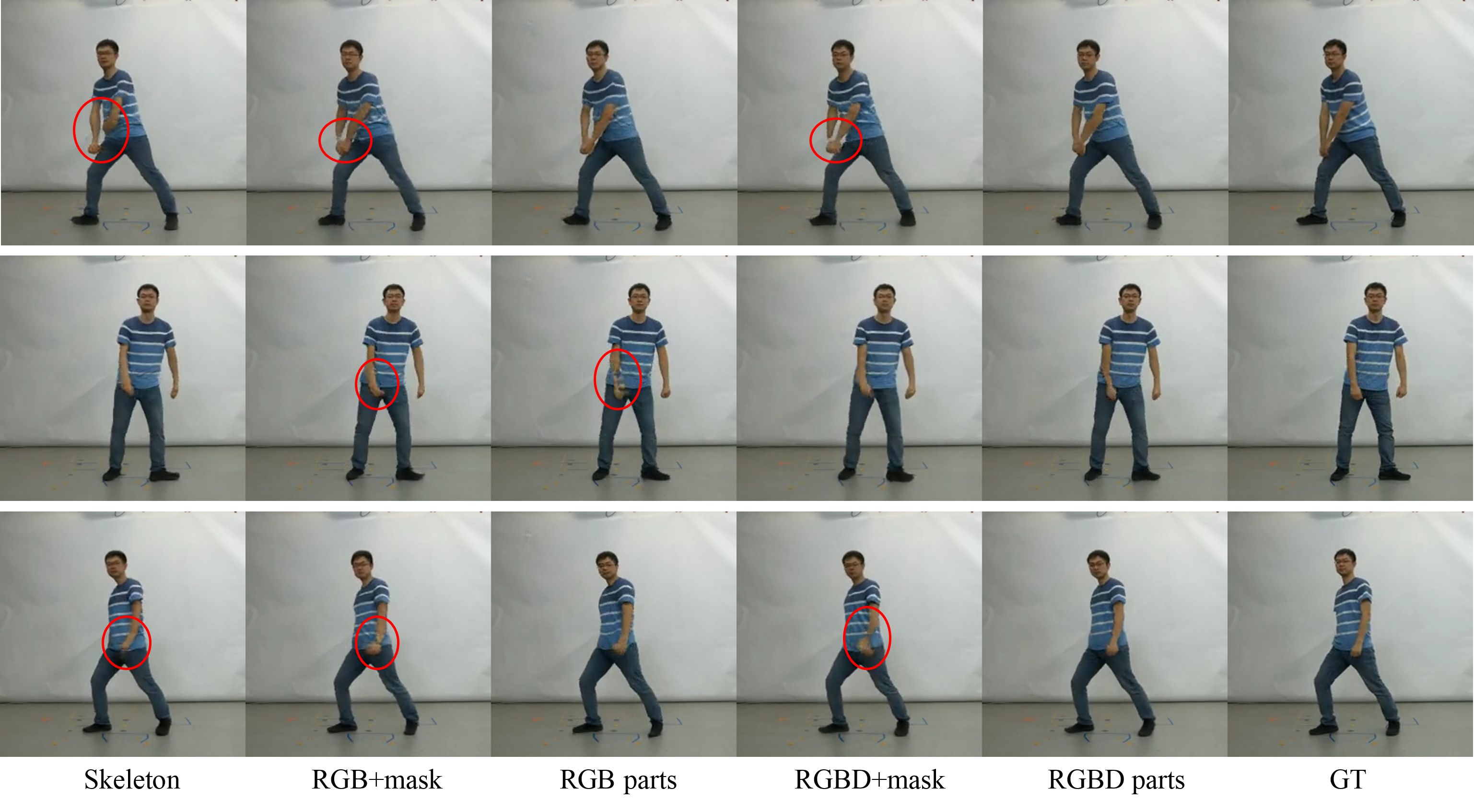}	
	
	\caption{
		Qualitative validation of different conditioning inputs.
	}
	\label{fig:ablation_input_qual}
\end{figure*}

Next, we evaluate our design choices and study the importance of each component of our proposed approach. The reported errors are always computed on the foreground only, which we determine based on background subtraction.

\paragraph{Conditioning Input:} First, we analyze the effect of different conditioning inputs. To this end, we compare the use of the following input modalities:
\begin{itemize}
\item[1)] rendered skeleton (\emph{skeleton}),
\item[2)] rendered RGB mesh and semantic masks (\emph{RGB+mask}),
\item[3)] per-body-part rendered mesh RGB only (\emph{RGB parts})
\item[4)] rendered mesh RGBD and semantic masks (\emph{RGBD+mask}),
\item[5)] per-body part rendered mesh RGBD (\emph{RGBD parts}, \textbf{ours}).
\end{itemize}

\begin{figure}[t!]
\includegraphics[width=\columnwidth]{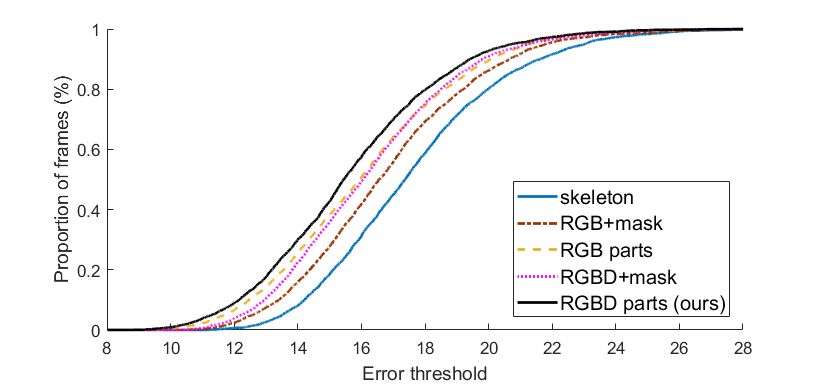}	
	\caption{
		Quantitative validation of different conditioning inputs.
		We show the proportion of frames on the vertical axis that have an error mean for the region of the person in the foreground smaller than the threshold on the horizontal axis. 
		Our full approach consistently obtains the best results.
	}
	\label{fig:ablation_input}
\end{figure}

In Figs.~\ref{fig:ablation_input} and~\ref{fig:ablation_input_qual} we show the quantitative and qualitative results, respectively, where it is revealed that
using the rendered RGB mesh in conjunction with semantic masks (\emph{RGB+mask}, red dashed-dotted line) is superior compared to using only a sparse skeleton for conditioning (\emph{skeleton}, solid blue line). Moreover,  explicitly applying the semantic masks to the rendered images (\emph{RGB parts}, dashed yellow line), i.e. breaking the image into its semantic parts, significantly improves the results.
The results with depth information \emph{RGBD+mask} (pink dotted line) and \emph{RGBD parts} (black line) are consistently better than the RGB-only results.
As can be also seen in Fig.~\ref{fig:ablation_input_qual} the depth information improves the image synthesis results in general.
Moreover, in frames where body-part occlusions exist the depth information helps to reduce the artifacts. 
We also find that using only the part-based rendered mesh (\emph{RGB parts}, dashed yellow line) in the conditioning input, without additional rendered depth images, is inferior to our final approach. 
We observe that the additional depth information (\emph{ours}, solid black line) improves the quality of the synthesized imagery, since the network is better able to resolve which body parts are in front of each other.
Hence, to achieve better robustness for the more difficult occlusion cases, we decided to use the depth channel in all other experiments.
Tab. ~\ref{tab:ablation_quant} also confirms the observations made above quantitatively in terms of both L2 error and SSIM comparison to ground truth. 
Note that there is a large improvement from the \emph{RGBD+mask} to \emph{Ours} (\emph{RGBD parts}), which shows that the part-based representation plays a critical role in the improvement.
The intuition behind this is that the separation of body parts can enforce the network to distinguish body parts in different ordinal layers, which helps to improve the results in the presence of occlusion. 
Comparing to semantic masks, the separation functionality of our part-based representation is more explicit. 
Comparing RGB parts with Ours (RGBD parts), we can see that, while the influence of the depth input to the visual quality in the example in the last row of Fig.~\ref{fig:ablation_input_qual} is relatively subtle, the depth information helps in general as indicated by the comparisons in both L2 error and SSIM metrics (see Table.~\ref{tab:ablation_quant}).
This shows that the part-based representation plays a critical role in the improvement.
	The intuition behind this is that the separation of body parts can enforce the network to distinguish body parts in different ordinal layers, which helps to improve the results in the presence of occlusion. The separation functionality of our part-based representation is more explicit than semantic masks. 
	This is supported by our experimental results both quantitatively and qualitatively.

\begin{figure}[t!]
\includegraphics[width=\columnwidth]{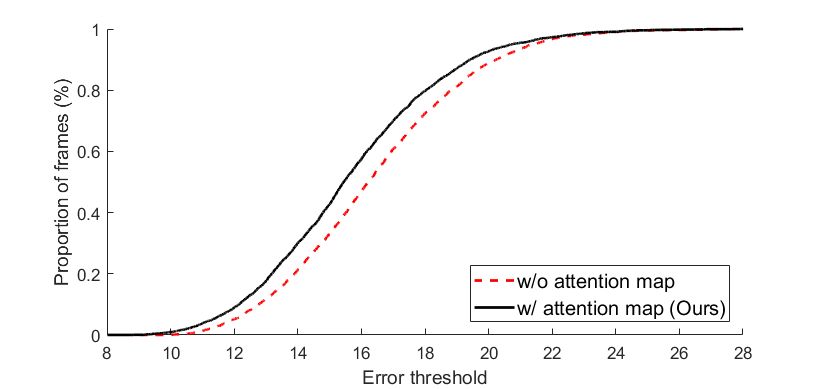}	
	\caption{
		Quantitative validation of attention map mechanism.
		We show the proportion of frames on the vertical axis that have an error mean for the region of the person in the foreground smaller than the threshold on the horizontal axis. 
		Consistently better results are obtained with our attention map.
	}
	\label{fig:ablation_attention}
\end{figure}

\begin{figure}[t!]
\includegraphics[width=\columnwidth]{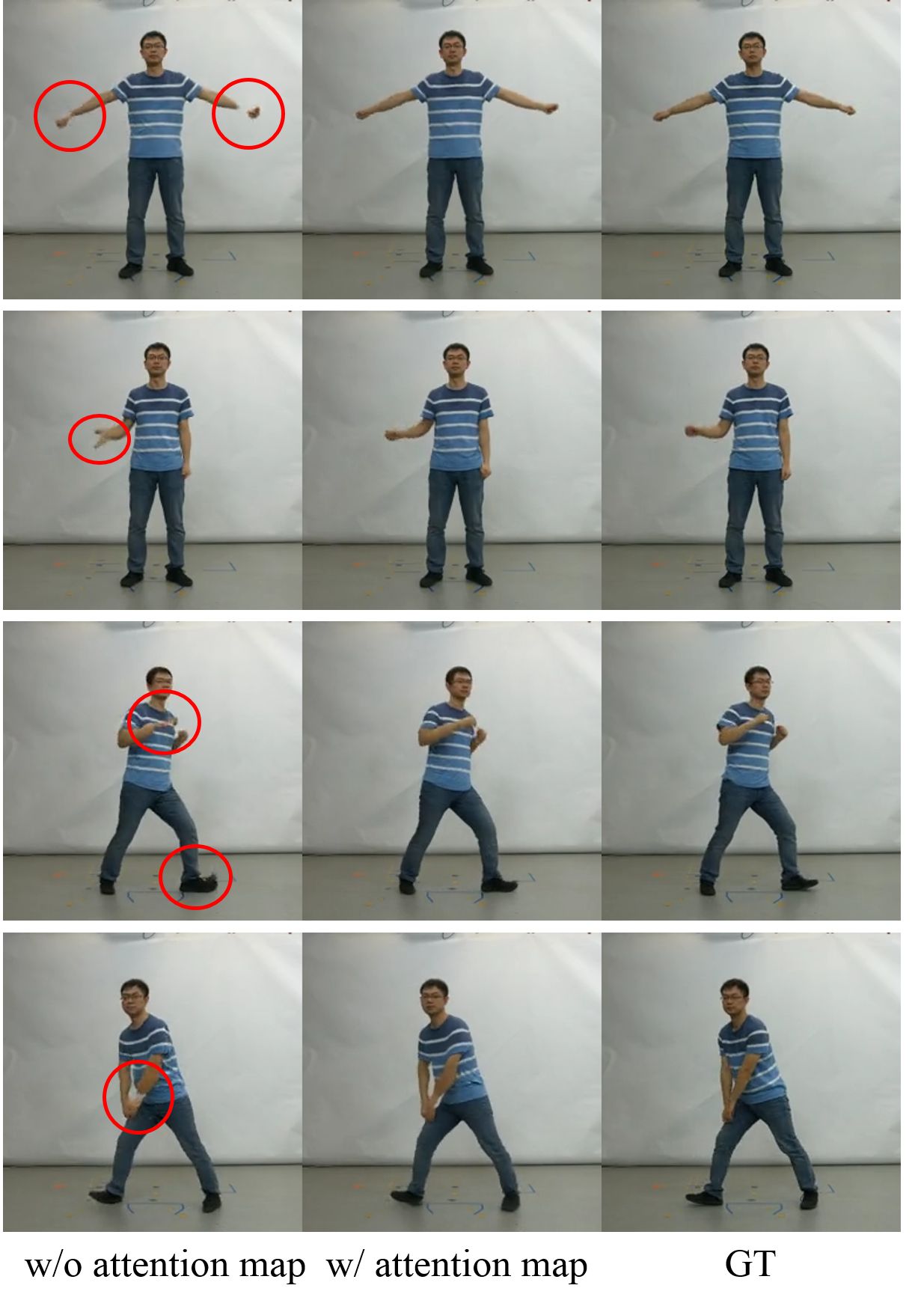}	\vspace{-0.6cm}
	\caption{
		Qualitative validation of attention map mechanism in our attentive discriminator. The comparison shows that our attention map based synthesis leads to a clear quality improvement (particularly visible in regions encircled in red), as seen by comparison to the ground truth.
	}
	\vspace{-0.5cm}
\label{fig:ablation_attention_qual}
\end{figure}

\paragraph{Attentive Discriminator:}
Moreover, we study the effect of using the proposed attention map mechanism in our attentive discriminator.
In Fig.~\ref{fig:ablation_attention}, it can be seen that using the attention GAN (solid black line) yields better results than a network trained without the attentive discriminator (dotted pink line). We show these improvements visually in Fig.~\ref{fig:ablation_attention_qual}.
We also confirm this observation quantitatively in terms of L2 and SSIM errors, see Tab.~\ref{tab:ablation_attention_quant}.

\begin{table}
	\centering
	\small
	\caption{Quantitative evaluation. We calculate the L2 error and SSIM for the region of the person in the foreground in each image and report the mean value for the whole sequence. Our full approach obtains the best scores.}
	\begin{tabular}{ | l | c | c |}
		\hline
		& \textbf{L2 error} & \textbf{SSIM} \\ \hline \hline
		\textbf{skeleton} & 17.64387 & 0.60176 \\ \hline
		\textbf{RGB+mask} & 16.82444 & 0.63102 \\ \hline
		\textbf{RGB part} & 16.12499 & 0.64023 \\ \hline
		\textbf{RGBD+mask} & 16.25977 & 0.64199 \\ \hline
		\textbf{Ours} & \textbf{15.67433} & \textbf{0.65328} \\ \hline
	\end{tabular}
	\label{tab:ablation_quant}
\end{table}

\begin{table}
\centering
\small
\caption{Quantitative evaluation of attentive discriminator. We calculate the L2 error and SSIM for the region of the person in the foreground in each image and report the mean value for the whole sequence. Our full approach obtains the best scores.}
	\begin{tabular}{ | l | c | c |}
		\hline
		  & \textbf{L2 error} & \textbf{SSIM} \\ \hline \hline
		\textbf{No attention} & 16.39865 & 0.64320 \\ \hline
		\textbf{Ours} & \textbf{15.67433} & \textbf{0.65328} \\ \hline
	\end{tabular}
	\label{tab:ablation_attention_quant}
\end{table}

\begin{figure}[t]
	\centering
	\includegraphics[width=0.93\columnwidth]{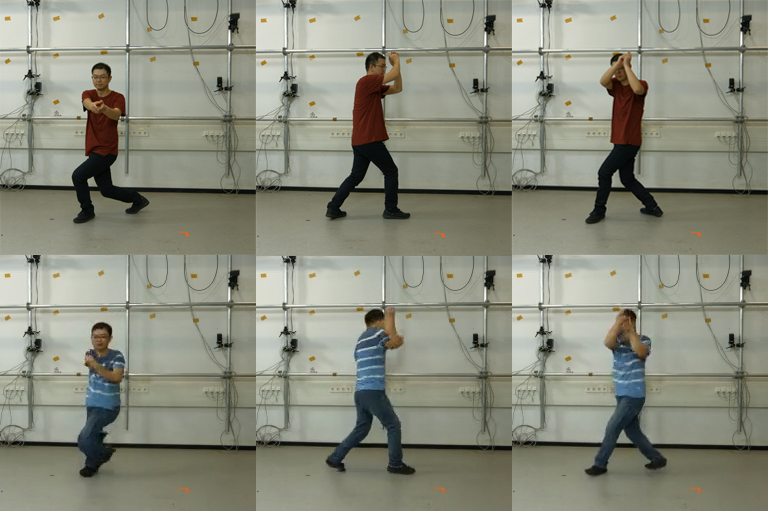}\\
	(a) \hspace{2cm}	(b) \hspace{2cm} 	(c) 
	\vspace{-0.3cm}
	\caption{
		Failure cases due to tracking error (a), rarely seen poses in the training corpus (b) and strong occlusion (c).
	}
	\label{fig:failure}
\end{figure}
%

\section{Limitations}
Despite the fact that we have presented compelling full body reenactment results for a wide range of motion settings, many challenges and open problems remain, which we hope will be addressed in the future.
Synthesizing severely articulated motions is very challenging for all kinds of learned generative models, including ours, due to multiple reasons: 
(i) articulated motions are highly non-linear,
(ii) self-occlusions in human performances introduce discontinuities,
(iii) monocular tracking imperfections degrade the quality of the training corpus, and
(iv) challenging poses are often underrepresented in the training data.
Fig.~\ref{fig:failure} shows several typical failure cases of our method.
Artifacts arise in particular at the end-effectors, e.g., hands or feet, since they undergo strong changes in spatial position and rotation.
One potential solution could be to split the network into different branches for each body part, possibly into further sub-branches depending on the pose or view-point, while jointly learning a differentiable composition strategy.

	Interactions with objects are challenging to synthesize for our as well as related techniques.
	This would require to jointly capture body pose as well as object position and shape at high accuracy, which is currently an unsolved problem.

	Occasional local high-frequency artifacts are due to the specific choice of the used GAN architecture. Completely removing such local patterns, which are often observed in outputs of GANS, remains an open challenge.

Even though our results exhibit high quality, a temporally coherent synthesis of human performances that is free of temporal aliasing is highly challenging. This is also due to the non-linearities of articulated motion, which is particularly noticeable for fine-scale texture details.

We have conducted experiments on incorporating temporal information by concatenating several adjacent frames as input to the network.
However, the results are not significantly better than with our proposed method.
We still believe that a more sophisticated integration of temporal information might further improve the results.

For example, a space-time adversarial consistency loss, which operates on a small time slice, could help to alleviate local temporal flickering.

Another possible solution are recurrent network architectures, such as RNNs or LSTMs.

Currently, our networks are trained in a person-specific manner based on a long training sequence.
Generalizing our approach, such that it works for arbitrary people given only a single reference image as input is an extremely challenging, but also very interesting direction for future work.

\section{Conclusion}
In this paper we have proposed a method for generating video-realistic animations of real humans under user control, without the need for a high-quality photorealistic 3D model of the human.
Our approach is based on a part-based conditional generative adversarial network with a novel attention mechanism.
The key idea is to translate computer graphics renderings of a medium-quality rigged model, which can be readily animated, into realistic imagery.
The required person-specific training corpus can be obtained based on monocular performance capture.
In our experiments, we have considered the
reenactment of other people, where we have demonstrated that our approach outperforms the state-of-the-art in image-based synthesis of humans.

We believe this is a first important step towards the efficient rendition of video-realistic characters under user control.
Having these capabilities is of high importance for computer games, visual effects, telepresence, and virtual and augmented reality.
Another important application area is the synthesis of large fully annotated training corpora for the training of camera-based perception algorithms.

\begin{acks}
We thank our reviewers for their invaluable comments. We also thank Liqian Ma for his great help with comparison; Franziska Mueller and Ikhsanul Habibie for data acquisition; Jiatao Gu for discussion. This work was supported by ERC Consolidator Grant 4DRepLy (770784), Max Planck Center for Visual Computing and Communications (MPC-VCC) and ITC of Hong Kong (ITS/457/17FP).
\end{acks}

\bibliographystyle{ACM-Reference-Format}
\bibliography{main}

\end{document}